\documentclass[letterpaper, 10 pt, conference]{ieeeconf}  % Comment this line out if you need a4paper

\IEEEoverridecommandlockouts                              % This command is only needed if 
\usepackage{cite}
\usepackage{graphicx}
\usepackage{color}
\newcounter{RomanNumber}

\newcommand{\MyRoman}[1]{\setcounter{RomanNumber}{#1}\Roman{RomanNumber}}

\title{\LARGE \bf
VH-HFCN based Parking Slot and Lane Markings Segmentation on Panoramic Surround View
}

\author{Yan Wu, Tao Yang, Junqiao Zhao$^{*}$, Linting Guan and Wei Jiang % <-this % stops a space
\thanks{This work is supported by the National Natural Science Foundation of China (No. U1764261), the Natural Science Foundation of Shanghai (No.kz170020173571) and the Fundamental Research Funds for the Central Universities (No. 22120170232, No. 22120180095).}% <-this % stops a space
\thanks{All authors are with the Department of Computer Science and Technology, School of Electronics and Information Engineering, Tongji University, Shanghai, 201804, China
        {\tt\small zhaojunqiao@tongji.edu.cn}}%
}

\begin{document}

\maketitle
\thispagestyle{empty}
\pagestyle{empty}

%%%%%%%%%%%%%%%%%%%%%%%%%%%%%%%%%%%%%%%%%%%%%%%%%%%%%%%%%%%%%%%%%%%%%%%%%%%%%%%%
\begin{abstract}

The automatic parking is being massively developed by car manufacturers and providers.
% The key to this system is the robust detection of parking slots and road structures, such as lane markings.
% Until now, robust parking slots and road structures detection method is the key issue to the automatic parking system.
Until now, there are two problems with the automatic parking. 
First, there is no openly-available segmentation labels of parking slot on panoramic surround view (PSV) dataset.
Second, how to detect parking slot and road structure robustly. 
Therefore, in this paper, we build up a public PSV dataset. 
At the same time, we proposed a highly fused convolutional network (HFCN) based segmentation method for parking slot and lane markings based on the PSV dataset.
% In this paper, we proposed an HFCN-based segmentation method for parking slot and lane markings in a panoramic surround view (PSV) dataset.
A surround-view image is made of four calibrated images captured from four fisheye cameras.
We collect and label more than 4,200 surround view images for this task, which contain various illuminated scenes of different types of parking slots.
A VH-HFCN network is proposed, which adopts an HFCN as the base, with an extra efficient VH-stage for better segmenting various markings.
The VH-stage consists of two independent linear convolution paths with vertical and horizontal convolution kernels respectively.
This modification enables the network to robustly and precisely extract linear features.
We evaluated our model on the PSV dataset and the results showed outstanding performance in ground markings segmentation.
Based on the segmented markings, parking slots and lanes are acquired by skeletonization, hough line transform and line arrangement.

\end{abstract}

%%%%%%%%%%%%%%%%%%%%%%%%%%%%%%%%%%%%%%%%%%%%%%%%%%%%%%%%%%%%%%%%%%%%%%%%%%%%%%%%
\section{INTRODUCTION}
An autonomous car is a vehicle that is capable of sensing its environment and driving without manual intervention. 
In the past decade, autonomous driving has entered a period of rapid development, especially in detecting surroundings with varieties of sensors, such as cameras, LiDAR.
% However, modern autonomous driving is mainly implemented in the urban path and fast express road.
However, as a branch of autonomous driving, efficient and highly automatic parking has still been a difficult problem.
There already exist many advanced parking assistance systems, but most of them are in a semi-manufactured stage or only suitable for a very limited parking lot.
The critical problem for automatic parking is the detection of parking slot and auxiliary lane markings.
Existing detection methods are mostly based on hand-crafted computer vision processing by extracting line features and adding artificial rules on them.
Parking slot and lane markings detected in this way are vulnerable to illumination change and fading, thus cannot be applied robustly and fully automatically.
%What is more, few cameras are used in early parking system, which would result in some blind spots.
%Therefore, modern detecting methods are based on panoramic surround view (PSV), which keep a full perspective on the vehicle.

In recent years, deep learning-based approaches have achieved great success in image recognition, object detection, and image segmentation \cite{Lecun2015Deep}.
The outstanding performance of deep learning makes it the preferred choice for sensing algorithm in autonomous driving.
Detecting parking slot and lane markings with deep network models have become a new direction.
However, the lack of openly-available parking slot labels based on PSV is the biggest obstacle.

In this paper, we build up a PSV dataset and propose a deep convolutional network to segment parking slot and lane markings. 
With the segmented ground markings, followed by post-processing of skeletonization, hough line transform, and line arrangement, we can robustly extract read-to-use parking slots and lanes for autonomous parking.
The main contributions of this paper are listed as follow:
\begin{itemize}
\item We collected more than 4,200 images of PSV, which contained scenes of various illumination conditions.
We labeled these images so that we could train a deep neural network for parking slot and lane markings segmentation.
\item We proposed an efficient and precise VH-HFCN model to segment the parking slot and lane markings around the vehicle on our PSV dataset. %The proposed VH-stage module is very efficient in linear feature extraction.
\end{itemize}

\section{RELATED WORK}

Automatic parking is an emerging industry in recent years.
In 2008, Yu-Chih Liu et al. staged a PSV to obtain the complete perspective around the car \cite{liu2008bird}. 
After then, Mengmeng Yu et al. used surround cameras to obtain a 360-degree surround view in the parking guide \cite{yu2014360}.
Recently, Chunxiang Wang \cite{wang2014automatic}, Jae Kyu Suhr \cite{suhr2014sensor}, Linshen LI \cite{li2017vision} used a PSV to realize the parking slot detection in automatic parking.
Apart from the parking slots, autonomous parking also requires robust LKA during planning, which can hardly be satisfied based on mid-to-far range front view cameras.
Traditional lane detection methods are mostly based on the line detection, line fitting for lane markings and multi-frame fusion methods \cite{hillel2014recent}.
Similarly, robust and generic lane detection methods are also required, especially in the surrounding of the vehicle.
In 2014, Jihun Kim used a convolutional neural network and random sampling method to acquire robust lanes \cite{kim2014robust}.
Subsequently, Baidu IDL proposed a deep convolutional network model for lane detection \cite{huval2015empirical}.
KAIST proposed VPGNet for detecting lanes and road signs with their labeled dataset \cite{lee2017vpgnet}.
However, these lane markings detection methods were all based on the images captured by the front monocular camera, which are only useful in the urban road.
% In this paper, lane markings detection is performed on PSV dataset. 
In 2016, Ford proposed to detect lane markings on both sides of the vehicle, but this approach has a limited camera perspective \cite{gurghian2016deeplanes}. 
% Thus it is still a new field to detect lane markings in surround view image.

Lane markings and parking slots present linear structures. Thus we propose to use semantic segmentation model to segment target parking slot and lane markings.
%Image semantic segmentation has achieved significant improvements in recent years.
Long, et al. proposed Fully Convolutional Network (FCN), in 2015. \cite{long2015fully}.
After then, deep networks based on this end-to-end learning methods have become the defacto of image semantic segmentation. 
Stacked Deconvolutional Network \cite{fu2017stacked} and SegNet \cite{badrinarayanan2015segnet} with a Encoder-Decoder Architecture performed well on public segmentation datasets.
% Most of the later network models are based on them.
Other improved methods include: using a more efficient benchmarking classification models, such as ResNet \cite{lin2016refinenet}, DenseNet \cite{jegou2017one}; optimizing network-specific modules like dilated convolution \cite{yu2015multi}, zoom-out features \cite{mostajabi2015feedforward}; combining the traditional segmentation model CRF \cite{chen2016deeplab}.
At CVPR 2017, Zhao et al. proposed a PSPNet model, achieving state-of-the-art performance on most public datasets \cite{zhao2016pyramid}.
Behrendt introduced a new lane markings segmentation network from automatically generated labels \cite{8202238}.

In prior work, based on FCN, we put forward efficient segmentation networks FCCN \cite{wu2017fully} and HFCN \cite{yang2018semantic}.
We proposed the cost function method to train our network.
% We further improved the up-sampling part and train with multiple cost functions.
Our approach was inspired initially by lane markings segmentation task, and we successfully evaluated it on public benchmarks.
In this paper, we made further improvements to implement a prominent network for parking slot and lane markings segmentation.

%TODO to be polished
\section{DATASET}

\subsection{Image Collection}

Current visual parking lot detection is mostly based on PSV, which has two main benefits: 
1) achieve a complete view around the car without blind spot; 
2) with a perspective of looking down, the parking slot and lane markings no longer distort and metric-based measurement becomes possible.
However, PSV datasets which can be used for deep learning are rare.
Therefore, this paper builds up a PSV dataset using TiEV (Tongji Intelligent Electronic Vehicle) based on roads in Jiading campus of Tongji University\footnote{http://cs1.tongji.edu.cn/tiev/resourse/}.

We collected a total of 4249 PSV images.
As to the distribution of training, validation, and test, we refer to other common datasets, by the ratio of 6: 1: 3.
In detail, the numbers of images for training, validation, and test are 2550, 425, and 1274.
Respectively, each part contains images captured in different scenes.

The original image of the surround view is captured by four cameras mounted in the front, rear, left, and right of the car body, each holds a resolution of 640x480.
All the cameras have a wide viewing angle of 180 degrees.
Because of heavy distortion, original images captured from the camera should be calibrated and undistorted.
% Thus we need to use internal parameters to realize fisheye correction operation first.
A car-centered coordinate frame will then be determined to warp the undistorted images onto the ground.
% of four cameras coordinate with warp perspective.
Fig. 1 shows the results of undistortion and warp perspective of the original image.

\begin{figure}[ht]
\centerline{\includegraphics[width=8.5cm]{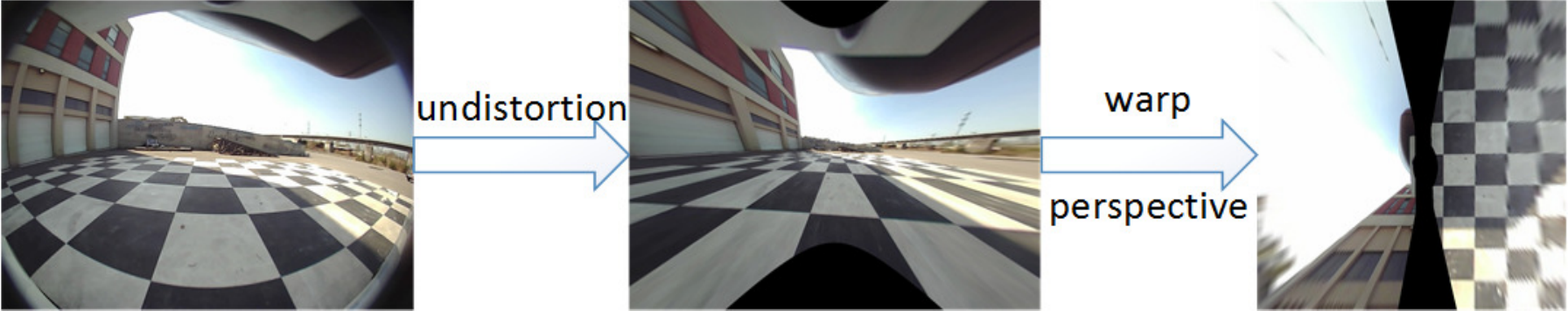}}
\label{figure1}
\vspace*{8pt}
\caption{Left image is the original image captured from the fisheye camera, the middle image is undistorted with internal parameters, the right image is the result of warp perspective transform.}
\end{figure}

After undistortion and warp perspective operation, we get four top-down view images that can be used for stitching.
We select four masks appropriately for each of the corresponding top-down views.
% It should be ensured that the four masks seamlessly stitched into a panoramic view.
Then, we stitch four top-down views with these masks to generate a PSV image.
Fig. 2 shows the four original images and the final PSV image, which holds a resolution of 1000 by 1000.
One pixel indicates 1 cm by 1 cm in car frame, so an area of 10 meters by 10 meters is covered.
% As a result, a PSV image can cover the surrounding scenes and ensure the image's clarity.

\begin{figure}[ht]
\centerline{\includegraphics[width=8.5cm]{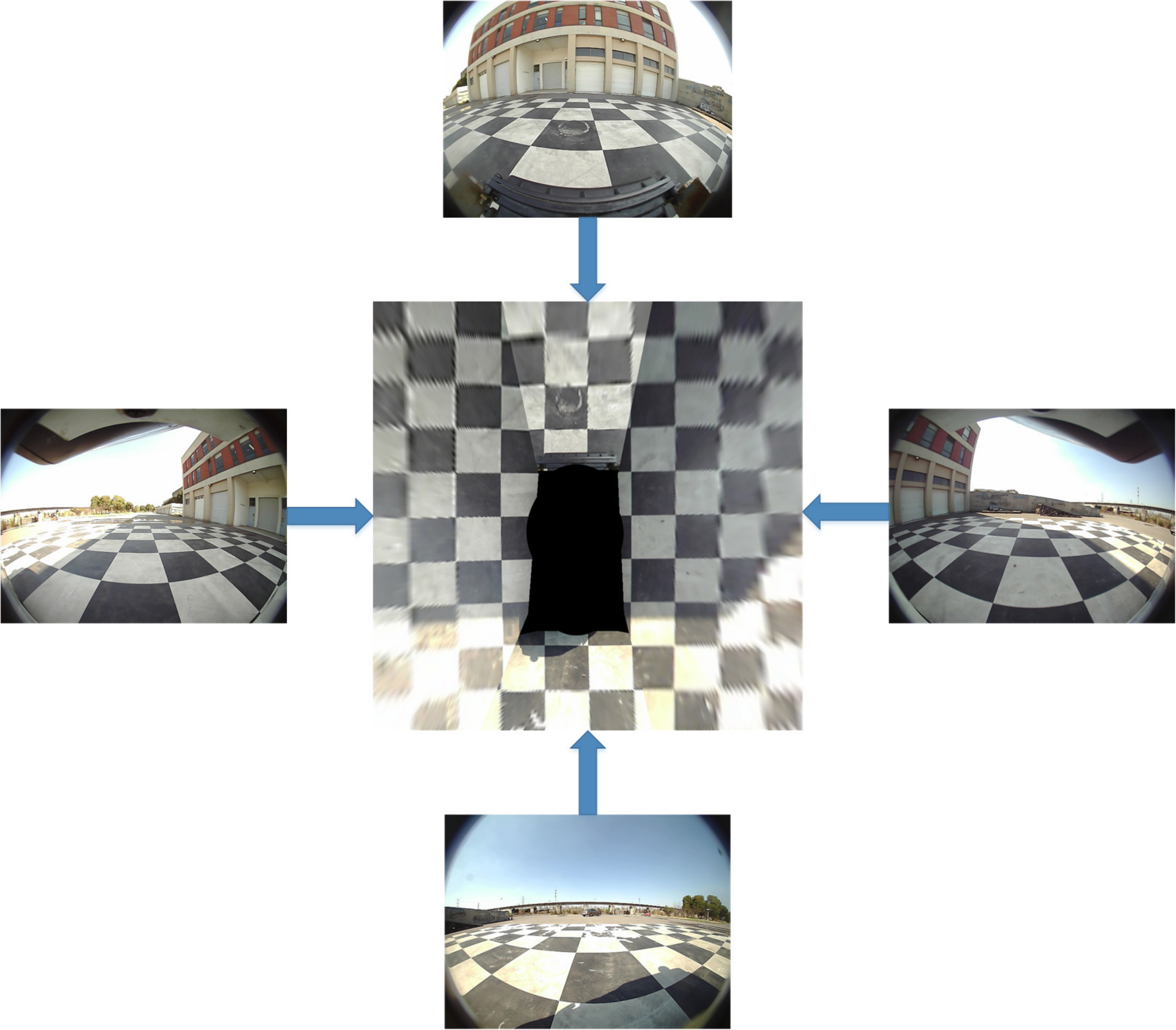}}
\label{figure2}
\vspace*{8pt}
\caption{A PSV image is generated by four original fisheye image.}
\end{figure}

We collected the PSV dataset containing a variety of surrounding environment, including indoor, outdoor, strong lighting, shadows and so on.
Both fuzzy and clear parking slots and lane markings were collected. 
For parking slots, we collected horizontal, vertical and diagonal shaped slots, also including different colors, like yellow and white.
Fig. 3 shows examples of different scenes and types.
Since lines hold only a small area in PSV images, the ratio of line segments targeted for identification is low.
We calculate a mean ratio 2.49\% of all images.

\begin{figure}[ht]
\centerline{\includegraphics[width=8.5cm]{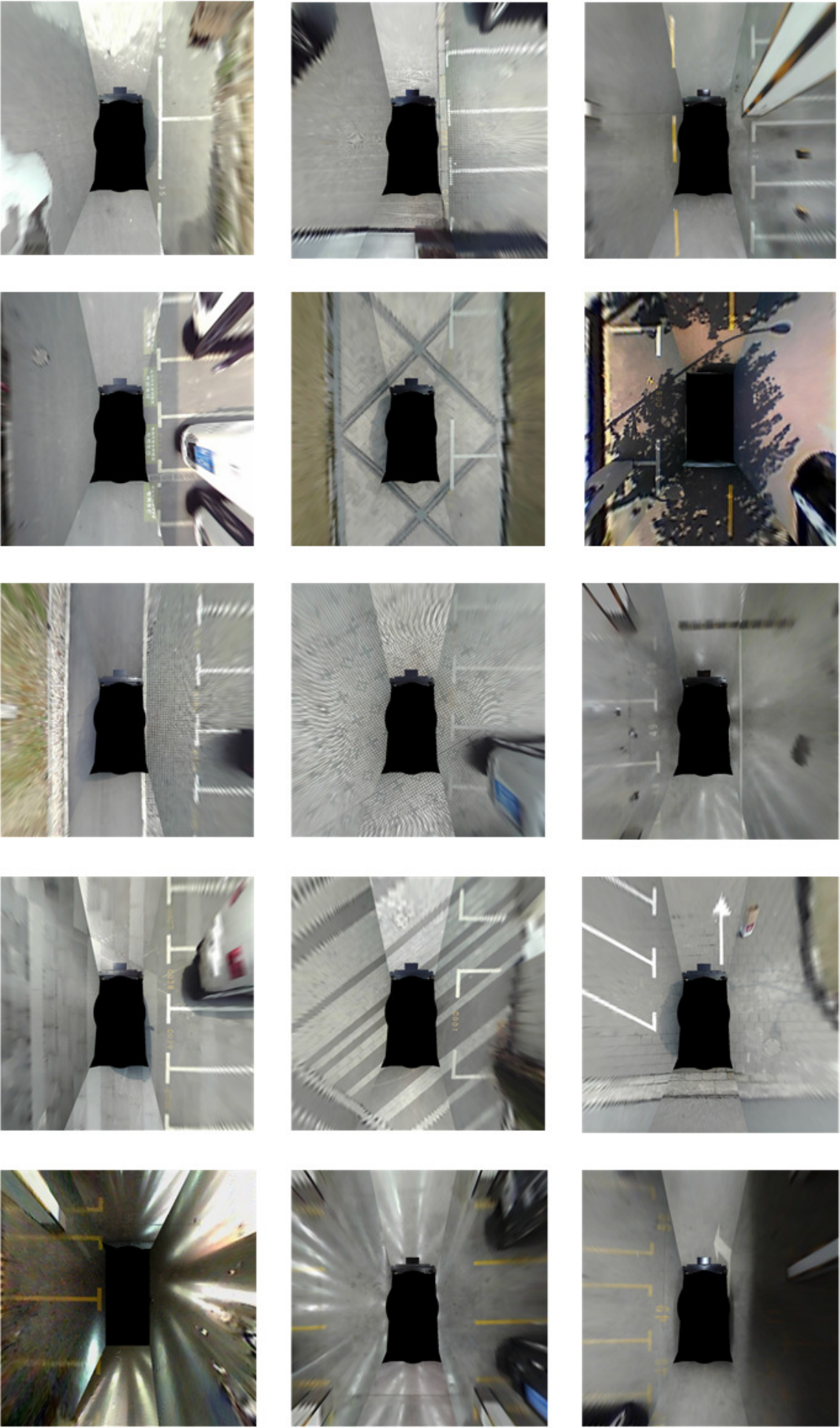}}
\label{figure3}
\vspace*{8pt}
\caption{Some sample images of our PSV dataset in different conditions: 1) first row contains outdoor and indoor scenes; 2) second row are under strong light and shadow; 3) third row shows both clear and fuzzy parking slots; 4) fourth row lists parking slots in horizontal, vertical and diagonal types; 5) fifth row are samples of yellow parking slots.}
\end{figure}

\subsection{Dataset Labeling}
There are five types of segmentation classes in our PSV dataset, which are parking slot, solid white lane, white dashed lane, yellow solid lane and yellow dashed lane.
When labeling the images, we connect all corner points of the parking slot or lane markings to generate a closed area as the segmentation target.
However, the labeling of the dashed lane is special.
We choose to label the dashed lane as a whole segmentation area, that is, the blank areas in the middle of the dashed lane are also labeled as a dashed lane area.
As shown in Fig. 4, the final labeled lane is an integral area, without appearing segmented.
The reason is to preserve the overall information of the lane, rather than segmenting the lane into disjointing blocks.
% \COMMENT{to be concise}
% According to the real effective area, a dashed lane only needs to label the white or yellow line segment that appears in the image.
% But if label dashed lane in this way, the overall information of the lane will be missed, and the difference between the dashed lane and solid lane will be slight.
% It is not easy for the network to differentiate solid lane and dashed lane.
% Therefore, 

\begin{figure}[ht]
\centerline{\includegraphics[width=8.5cm]{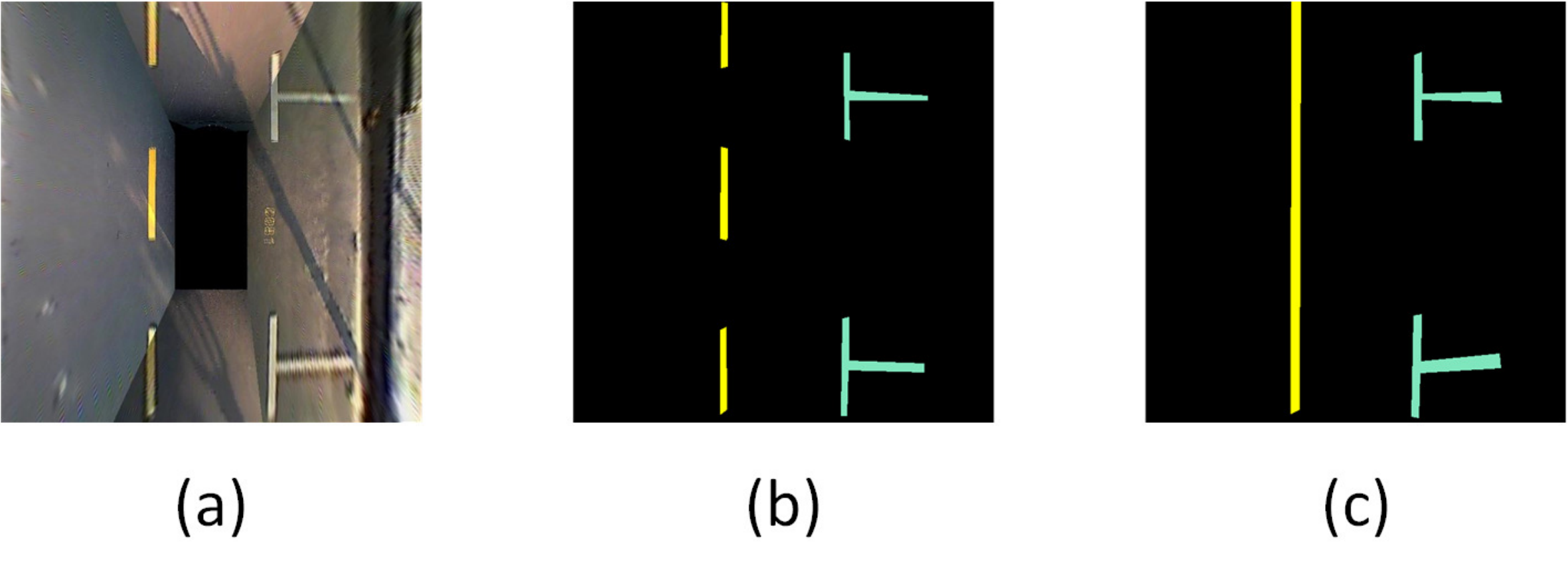}}
\label{figure4}
\vspace*{8pt}
\caption{The dashed line in (a) is labeled as an integral area like (c), rather than several segmented areas like (b).}
\end{figure}

\section{METHOD}

\subsection{Network}

We use a deep convolutional network to segment ground markings on our PSV dataset.
The proposed VH-HFCN model is illustrated in Fig. 5.
Our network is based on HFCN \cite{yang2018semantic}, which improves the up-sampling process of the FCN model, divide the up-sampling operation on the feature maps into five steps.
After each step, the feature maps expand to double size in both width and height, as a reverse operation of the previous pooling layer.
We then make a combination of each pair of up-sampled and down-sampled feature maps, so as to realize the fusion and reuse of low-level feature information.
In addition to this up-sampling path, we add an extra one-step up-sampling operation on each up-sampling layer and acquire a pre-output after each one-step up-sampling. 
As there exist five up-sampling layers, we shall acquire five pre-output. 
Finally, we concatenate these pre-outputs, followed by a convolution layer to generate the final output.

\begin{figure*}[ht]
\centerline{\includegraphics[width=18cm]{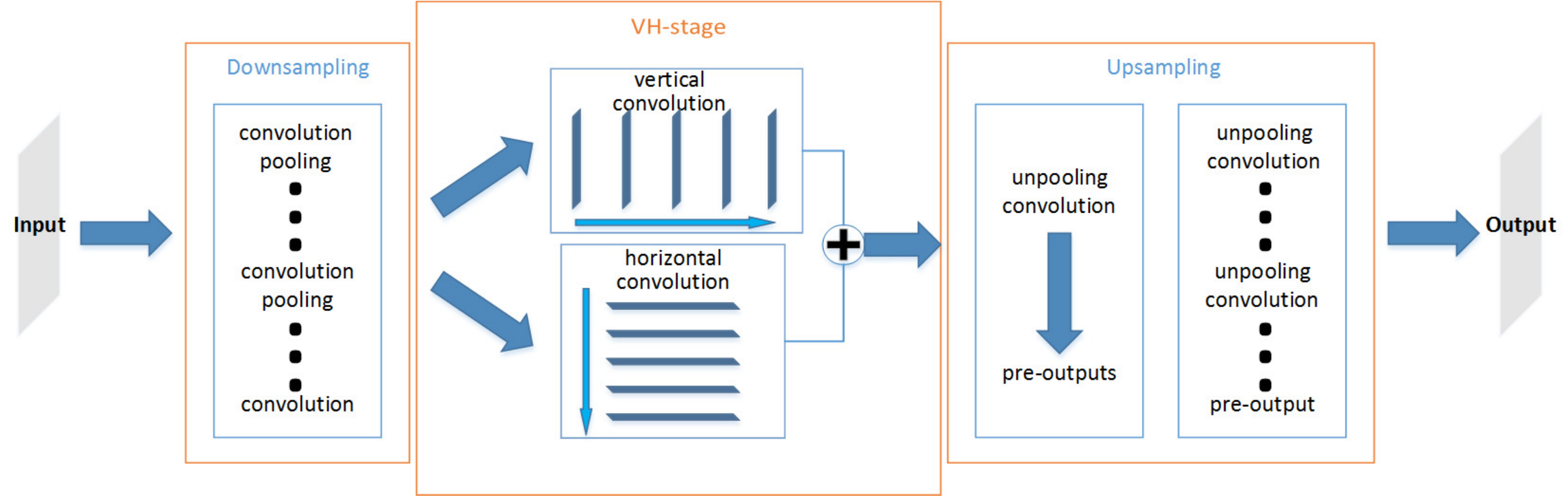}}
\label{figure5}
\vspace*{8pt}
\caption{Our segmentation network consists of three parts: the first down-sampling part for feature extraction; the second VH-stage part for further linear feature extraction, including independent horizontal and vertical convolution path; the last up-sampling part for feature fusion and resolution recover, containing two separate up-sampling way.}
\end{figure*}

% \COMMENT{Can you weaken this assumption?}

% We focus on studying the segmentation of parking slot and lane markings in this paper.
% It is easy for us to find that there is a distinct characteristic in these ground marks, that is, they both keep horizontal and vertical linear structures.
In most case, lane markings appear as vertical lines, or with a small angle of deflection, while parking slot is pairs of perpendicular lines.
Based on this fact, we add an extra VH-stage before the up-sampling operation to further learn these linear features.
This extra stage hardly destroys feature information learned in low layers. Thus other structures like slanting lines can also be well segmented.
The VH-stage consists of two independent special convolution paths; each contains five convolution layers and Relu layers.
The first path uses a series of vertical convolution kernels with a size of 9x1.
These special convolution kernels are of great help for extracting vertical linear features.
Similarly, horizontal convolution kernels with a size of 1x9 are helpful for extracting horizontal linear features.
We then sum these two paths and set the result as input to the subsequent up-sampling process.
By adding this extra VH-stage, we improved the segmentation performance, assessed by metric of mean Intersection-of-Union(mIoU), of the network by more than five percentage points.
We will describe this enhancement in more detail in next section. 

\subsection{Cost Function}
% In network training, we use the soft cost function we proposed in our prior work.
% The inspiration for this cost function comes from the using of FCN to achieve lane markings segmentation in offline perception competition of the future challenge of China.
In the offline perception competition of the future challenge in China, we adopt the segmentation network FCN to detect lanes, since traditional methods have difficulty in classifying all lane types. 
However, the network's learning of the target will be weakened due to the small proportion of pixels for the parking slots and lane markings in the image.
%,  due to the large proportion of the background in calculating the cost function.
Therefore, when calculating the cost function, we increase the weights of the target area.
What is more, the weights given to the target are not fixed.
We dynamically calculate the proportion of each target category in the current image and take the inverse of the proportion as the cost weight.
This cost function allows the network to focus more on the learning of the target area.
In addition, we add cost functions in five pre-outputs, which is introduced in section IV.A. 
Thus there are six cost functions when training our segmentation model.
The final Loss is calculated according to equation (1), in which $L_{final output}$ is the loss of final output and $L_{preoutputs_i}$ is the loss of $i_{th}$ pre output.
$\lambda_{i}$ is the weight of each loss, in this paper, $\lambda_{i}$ is we achieve the best performance by setting all $\lambda_{i}$ to 1.0.
With this additional loss, our training can reduce the error erosion during backpropagation.

\begin{equation}
Loss=L_{final output} + \sum{\lambda_{i}*L_{preoutputs_i}}
\label{equ1}
\end{equation}

Each single loss $L$ is acquired by equation (2), where $H_{i,j}$, $Y_{i,j}$ are predicted class and label in pixel(i, j), $w_{i,j}$ is class weight.
Additionally, $w_{i,j}$ is calculated along with the proportion of the target class in the whole image \cite{wu2017fully}.

\begin{equation}
L=\sum_{i=1}^m\sum_{j=1}^n{w_{i,j} \left \| H_{i,j}-Y_{i,j} \right \| ^2}
\label{equ2}
\end{equation}

\subsection{Extraction of Parking Slot and Lanes}

Though the VH-HFCN model can acquire segmented ground markings, further processing for getting the ready-to-use parking slots and lanes is necessary.
In this paper, post-processing after the segmentation consists of three steps: 1) apply morphological skeletonization to extract the central path of each segmented area; 2) hough line transform on the skeleton is added to generate necessary lines; 3) partition and merge similar lines, build up the parking slots and lanes.
After the processing, detected parking slots and lanes are used in automatic parking.
Though our extraction method is quite robust, it can be improved.
We will try continuous model fitting to realize the extraction in our future work.

\section{EXPERIMENTS}

\subsection{Experiments Configuration}

All our experiments are implemented on MatConvNet \cite{Vedaldi2015MatConvNet}, which is a Matlab toolbox for the deep convolutional network.
We used a batch-size of 10, a learning rate of 0.0001, and an epoch of 50 for training.
Multiple cost functions used in the experiments all keep a weight of 1.0.
% We also tried other weight proportions in HFCN \cite{yang2018semantic}, but get the best performance when all weights are set to 1.0.
The networks are trained on NVIDIA GeForce GTX TITAN X graphic card.

\subsection{Evaluation}

Our proposed model evaluation experiments on the PSV dataset can be divided into three stages: 1) acquire the best size of vertical and horizontal convolution kernels; 2) determine the convolution path and the method of combining feature maps in VH-stage; 3) compare segmentation performance of proposed model with FCN, FCCN, and HFCN on PSV dataset.
We use Mean IoU as the metric of our segmentation performance.

It is a crucial issue for us to determine the convolution kernel size in VH-stage, since short kernels are not enough to cover full linear features, while long kernels may reduce the efficiency of the network, even reduce the segmentation performance.
Table \MyRoman{1} shows the segmentation results of different convolution kernel sizes in VH-stage.
Among the model column, model v3h3 means horizontal kernels with a size of 1x3 and vertical kernels with a size of 3x1; the other models are similar to this.
From Table \MyRoman{1}, we can know that the longer the convolution kernel, the better the segmentation results.
However, if we continue to enlarge the kernel after model v9h9, the growth of the result has become negligible, the major metric mIoU even dropped.
Hence, Table \MyRoman{1} shows that the model v9h9 obtains the best performance.
It should be oversized convolution kernel with redundant feature information.
Therefore, there is no need for us to continue enlarging convolution kernel.

\begin{table}[h]
\caption{Results of different convolution kernel sizes in VH-stage (Mean IoU is the performance metric).}
\label{table1}
\scriptsize
\begin{center}
\renewcommand\arraystretch{1.5}
\begin{tabular}{|c||ccc|}
\hline
Model & Mean pix.acc(\%) & Pixel acc(\%) & Mean IoU(\%)\\
\hline
v3h3	&89.71		&95.30		&43.80 \\
v5h5	&89.59		&95.29		&44.75 \\
v7h7 	&88.25		&95.32		&45.72 \\
v9h9 	&89.79		&96.25		&\textbf{46.51} \\
v11h11 	&89.87		&95.88		&46.48 \\
\hline
\end{tabular}
\end{center}
\end{table}
For the methods of feature maps combination after the VH-stage, we tried both concatenation and summing.
Besides, we added a third independent convolution path apart from the vertical and horizontal convolution path.
Since there exist some slanting lanes in our PSV dataset, we add the third convolution path to retain feature information extracted in lower layers.
As is shown in Table \MyRoman{2}, the third convolution path is included in the first model conv+, while only vertical and horizontal path in model concat and sum.
We can know that without an additional convolution path we achieve a higher segmentation performance in mIoU, which shows that the VH-stage can learn useful line features.
The additional convolution path may result in redundant features. 
Hence, the segmentation results become worse.

In addition, we can also learn from Table \MyRoman{2} that summing the feature maps achieves a better performance rather than concatenating them after VH-stage.
Two feature maps can be integrated into one feature map with a summing operation, so the summed map can hold integral features of target ground markings, especially parking slot.
In other way, features are still separated by a concatenation, which results in a worse performance in later segmentation.

\begin{table}[h]
\caption{Results of different strategies in VH-stage (Mean IoU is the performance metric).}
\label{table2}
\scriptsize
\begin{center}
\renewcommand\arraystretch{1.5}
\begin{tabular}{|c||ccc|}
\hline
Model & Mean pix.acc(\%) & Pixel acc(\%) & Mean IoU(\%)\\
\hline
conv+	& 90.96		& 96.09		& 45.27 \\
concat 	& 88.27		& 96.29		& 45.99 \\
sum 	& 89.79		& 96.25		& \textbf{46.51} \\
\hline
\end{tabular}
\end{center}
\end{table}

Apart from our proposed network in this paper, we also evaluate FCN, FCCN, and HFCN on our PSV dataset.
Table \MyRoman{3} shows detail results of these models.
Compared with other models, we achieve a considerable increase in the segmentation performance, which indicates that our proposed VH-stage strategy is effective.
Fig. 6 shows several visual samples with predictions of different models on PSV dataset.
We can learn that our proposed model achieves more precise segmentation result.
It's worth noting that we achieve the almost perfect dashed lane segmentation in the fourth row, while other models generate bumps in the blank area.
This is due to the integral lane features obtained by VH-stage module.
As shown in Fig. 7, post-processing is added to the segmented result to generate the ready-to-use parking slots and lanes.

\begin{table*}[ht]
\centering
\caption{Quantitative results of segmentation performance on our PSV dataset (\%).}
\label{table3}
\renewcommand\arraystretch{1.5}
\begin{center}
\begin{tabular}{|c||cccccccc|}
\hline
Model & background & parking & white-solid & white-dashed & yellow-solid & yellow-dashed & pacc & mIoU   \\
\hline
FCN \cite{long2015fully}  & 85.88 & 13.16	& 18.42	& 7.40	& 23.09	& 20.32	& 86.18 & 28.04 \\
FCCN \cite{wu2017fully} & 92.53 & 22.50 & 29.60 & 11.87 & 41.21 & 27.38 & 92.66 & 37.51 \\
HFCN \cite{yang2018semantic} & 93.87 & 25.46	& 36.26	& 18.97	& 45.08	& 26.87	& 93.97 & 41.09 \\
Ours  & \textbf{96.22} & \textbf{36.16}	& \textbf{39.56}	& \textbf{21.46}	& \textbf{47.64}	& \textbf{38.03}	& \textbf{96.25}	& \textbf{46.51} \\
\hline
\end{tabular}
\end{center}
\end{table*}

\begin{figure*}[ht]
\centerline{\includegraphics[width=15.5cm]{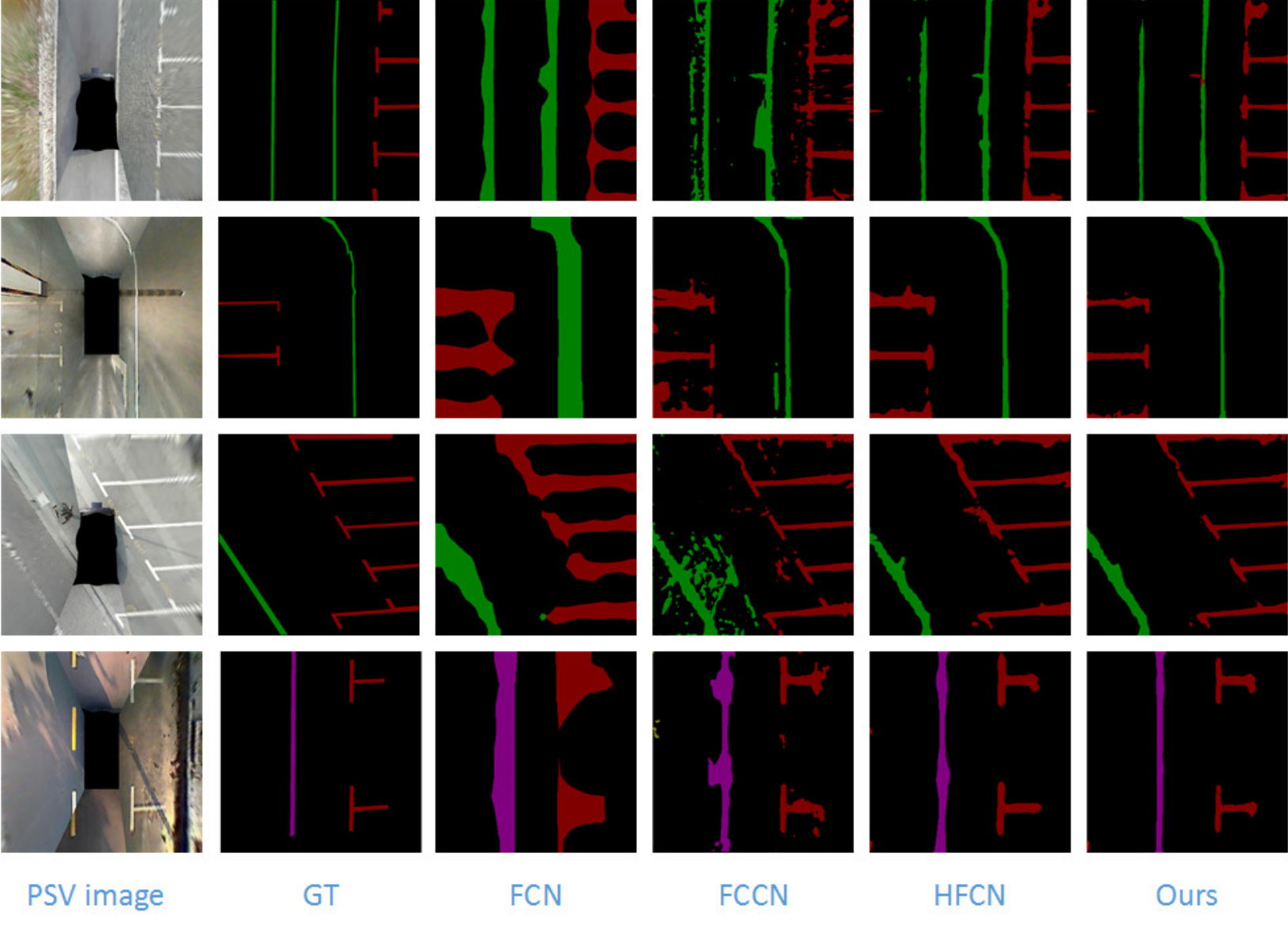}}
\label{figure6}
% \vspace*{8pt}
\caption{Some visual results of different models on PSV dataset. We achieve more precise segmentation on parking slot and lane markings.}
\end{figure*}

\begin{figure}[ht]
\centerline{\includegraphics[width=6.4cm]{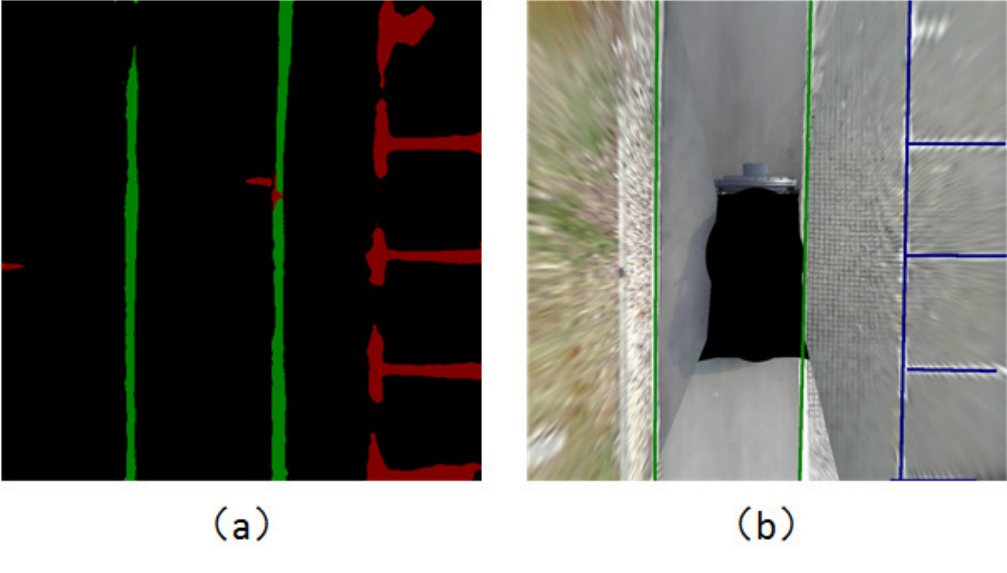}}
\label{figure7}
% \vspace*{8pt}
\caption{Parking slots and lanes detection based on segmentation result. (a) segmentation result by VH-HFCN; (b) detected lanes and space parkings.}
\end{figure}

\section{CONCLUSIONS}
In this paper, we introduce a parking slot and lane markings PSV dataset.
This dataset comprises surround-view images in varieties of scenes, with multiple types of the parking slot.
Based on the labeled PSV dataset, we propose a VH-HFCN segmentation model, which is proved to be efficient in extracting linear feature information from parking slot and lane markings.
We use multiple soft cost function to train our model, and achieve a significant performance in the PSV dataset.
Compared with other network models, we acquire a significant increase in performance.
In addition, we add post-processing to detect the final parking slots and lanes.
In the future, we intend to label more PSV images and design a further extended network to improve segmentation performance.
% Since there is still much room for improvement in our segmentation results, we will further improve our network performance in future work.

\bibliographystyle{IEEEtran}
\bibliography{IEEEabrv,ref}

\end{document}